\documentclass[letterpaper]{article} 
\usepackage{aaai24}  
\usepackage{times}  
\usepackage{helvet}  
\usepackage{courier}  
\usepackage[hyphens]{url}  
\usepackage{graphicx} 
\usepackage{natbib}  
\usepackage{caption} 
\usepackage{amssymb,amsmath,array}
\usepackage{algorithm}
\usepackage[noend]{algorithmic}
\usepackage{newfloat}
\usepackage{listings}
\DeclareCaptionStyle{ruled}{labelfont=normalfont,labelsep=colon,strut=off} 
\floatstyle{ruled}
\newfloat{listing}{tb}{lst}{}
\floatname{listing}{Listing}
\nocopyright

\title{Convolutional Channel-wise Competitive Learning for the Forward-Forward Algorithm}

\author {
    Andreas Papachristodoulou,
    Christos Kyrkou,
    Stelios Timotheou,
    Theocharis Theocharides
}
\affiliations {
    KIOS Research and Innovation Center of Excellence,\\
    Department of Electrical and Computer Engineering, University of Cyprus\\
    \{papachristodoulou.andreas, kyrkou.christos, stimo, ttheocharides\}@ucy.ac.cy
}

\begin{document}

\maketitle

\begin{abstract}
The Forward-Forward (FF) Algorithm has been recently proposed to alleviate the issues of backpropagation (BP) commonly used to train deep neural networks. However, its current formulation exhibits limitations such as the generation of negative data, slower convergence, and inadequate performance on complex tasks. In this paper we take the main ideas of FF and improve them by leveraging channel-wise competitive learning in the context of convolutional neural networks for image classification tasks. A layer-wise loss function is introduced that promotes competitive learning and eliminates the need for negative data construction. To enhance both the learning of compositional features and feature space partitioning, a channel-wise feature separator and extractor block is proposed that complements the competitive learning process. Our method outperforms recent FF-based models on image classification tasks, achieving testing errors of 0.58\%, 7.69\%, 21.89\%, and 48.77\% on MNIST, Fashion-MNIST, CIFAR-10 and CIFAR-100 respectively. Our approach bridges the performance gap between FF learning and BP methods, indicating the potential of our proposed approach to learn useful representations in a layer-wise modular fashion, enabling more efficient and flexible learning. Our source code is available at:

\url{https://github.com/andreaspapac/CwComp} .
\end{abstract}

\section{Introduction}

Artificial Neural Networks (ANNs) have emerged as a ubiquitous tool for tackling a variety of machine intelligence tasks. One of the most popular training algorithms for ANNs, Backpropagation (BP), employs the chain rule of differentiation for backward propagation of error gradients, applicable to any differentiable loss function. With its flexibility to accommodate complex architectures involving multiple layers and non-linear activation functions, BP can manage large datasets and high-dimensional input spaces.

However, BP is not without its limitations. Its assumption of perfect knowledge of forward pass computations contributes to the 'black-box' nature of models, challenging interpretability and transparency. Additionally, its task-specific feature learning limits model generalizability, requiring extensive labeled data and increasing susceptibility to adversarial attacks \cite{moraitis2022softhebb, madry2018deep}. Deep neural networks, inherently susceptible to issues like local minima and gradient problems due to non-convexity, face exacerbated challenges with BP's opacity, prompting the need for additional mitigation strategies \cite{lawrence2000overfitting, NIPS2000_059fdcd9, NEURIPS2022_dc06d4d2}. The limitations of BP have spurred research into more biologically plausible and hardware-efficient alternatives \cite{shinozaki2020biologicallymotivated}.

To circumvent these constraints, several non-BP training techniques have been proposed, including Predictive Coding \cite{rao1999predictive, salvatori2022reverse}, Evolutionary Algorithms \cite{Salimans2017}, non-BP Reinforcement Learning (RL) \cite{sutton2018reinforcement}, and Competitive Hebbian Learning \cite{miconi2021hebbian}. While these strategies demonstrate better handling of black-box models and vanishing/exploding gradients, they are often computationally demanding, require substantial memory, and present other limitations. A recent development in this domain is the Forward-Forward algorithm (FF) \cite{hinton2022forwardforward}, a non-BP technique involving two forward passes - one with positive (real) data, and the other with negative data. FF offers notable computational benefits in scenarios where the precise details of the forward computation are not known without the necessity of storing neural activities or interrupting for error derivative propagation. It is particularly advantageous for applications involving low-power analog hardware. Despite its advantages, the FF algorithm also exhibits slow convergence, inadequate performance on complex tasks, and the necessity of generating negative data.

This paper presents an approach to enhance the Forward-Forward Algorithm and overcome its shortcomings by concurrently extracting and separating features through Convolutional Neural Networks (CNN). Our framework results in improved representation learning and expedited training. Specifically, our main contributions are the following:
\begin{itemize}
\item Construct a layer-wise loss function that exploits channel-wise competitive learning. It reformulates the goodness function to eliminate the need for negative data construction and empowers each CNN layer to function as a standalone classifier.
\item Introduce the Channel-wise Feature Separator and Extractor (CFSE) block that partitions the feature space via channel-wise grouped convolutional layers and simultaneously enables the network to learn compositional features via standard non-separable convolutional layers.
\end{itemize}

Our approach is benchmarked against FF-based and non-Backpropagation methods reported in the literature on image classification tasks. Our method outperforms all FF-based models on basic image classification tasks, demonstrating quicker convergence rates and significantly better performance, achieving testing errors of 0.58\%, 7.69\%, 21.89\%, and 48.77\% on MNIST \cite{lecun1998gradient}, Fashion-MNIST \cite{xiao2017fashionmnist}, CIFAR-10, and CIFAR-100 \cite{krizhevsky2009learning} respectively. Moreover, it significantly outperforms almost all the non-BP models, with SoftHebb \cite{journé2022hebbian} being the only competitive one.

\section{Background and Related Work}

The BP algorithm has seen numerous improvements in the literature. Techniques like batch normalization \cite{ioffe2015batch}, dropout, and weight initialization have been employed to mitigate issues such as overfitting, vanishing gradients, and initialization bias. Moreover, alternative optimization methods such as Adam \cite{kingma2015adam} and Adagrad \cite{duchi2011adaptive} demonstrate improved convergence and generalization. Methods that address symmetric forward and backward weights, waiting times, and reliance on expensive-to-generate labeled datasets are explored to tackle the biological implausibility and hardware inefficiency of BP \cite{Sze2017EfficientPO}. Self-supervised learning schemes, for instance, adjust network weights through self-generated supervisory errors propagated from the system's output backward in an end-to-end manner \cite{chen2020simple}, though these are constrained by the efficiency of distributed computing hardware.

In response to the limitations of traditional deep learning algorithms based on BP, various non-BP training schemes have been proposed in the literature. Predictive Coding operates on the principle of the brain's predictive model, which continuously generates predictions and compares them with incoming sensory data \cite{NEURIPS2020_5cd5058b, ororbia2023predictive}. Similarly, Evolutionary Algorithms \cite{Salimans2017} adopt the concept of natural selection, with the fittest individuals (or weights in the context of ANNs) selected for reproduction. Both techniques, unlike BP, are not end-to-end training processes, and are better in managing black-box models and the vanishing/exploding gradients problem compared to BP, despite having no perfect knowledge of forward pass computation \cite{NEURIPS2020_e8d66338}. However, these methods are computationally demanding, memory-intensive, and, in the case of Predictive Coding, struggle with noise handling \cite{rao1999predictive, ofner2022generalized}. A form of Reinforcement Learning can manage black-box models by correlating random perturbations to weights or neural activities with changes in a payoff function. However, its high variance and computational expense make it difficult to scale to larger networks \cite{hinton2022forwardforward, sutton2018reinforcement}.

Feedback Alignment (FA) \cite{lillicrap2014random} offers a notable alternative to traditional backpropagation. Instead of fixed, symmetrical weights between forward and backward passes, FA uses random initial feedback weights that the network learns to optimize. This enables independent training of hidden layers, starting from zero initial conditions. FA contributes to biologically plausible models by using mostly local error signals and eliminating the need for symmetrical weights. Its extension, Direct Feedback Alignment \cite{nokland2016direct, Frenkel_2021}, further simplifies weight updates by avoiding non-local computations. Competitive Hebbian Learning enables neurons to compete for activation, enhancing the ones most responsive to input \cite{miconi2021hebbian, NIPS2008_07cdfd23}. While effective in overcoming overfitting and task-specific constraints, it suffers from slow convergence and parameter tuning with SoftHebb \cite{journé2022hebbian} promising higher convergence rates and accuracy by training deep networks without feedback signals.

The FF algorithm \cite{hinton2022forwardforward} has also emerged as a compelling alternative to the traditional BP method. It uses two forward passes, one with positive (real) data and the other with negative data, i.e., fake data generated by the network. One approach for generating a positive-negative pair involves the creation of a vector that combines the correct labels and corresponding data for positive examples while generating "bad data" by pairing input images with incorrect labels. The primary objective of each network layer under this framework is to maximize the model's performance on positive data and minimize it on negative data. This is achieved by optimizing a proposed "goodness" function, initially defined as the sum of squared activations within a layer. The layer-wise training structure of the FF algorithm enables the evaluation of each layer’s performance and facilitates a more transparent selection of hyperparameters and model architecture. Furthermore, the FF algorithm can operate without knowledge of forward computation details or the need for storing neural activities or halting error derivatives. The FF algorithm presents three main challenges. First, the need for negative data is highly task and dataset-specific. Second, the initial formulation of the goodness function is preliminary and calls for further refinement. Third, the FF algorithm is slower than BP and does not generalize as well on certain straightforward problems, which in practice renders BP more suitable for more complex models and datasets. Recent FF-based approaches have emerged that try to overcome the limitations of the original FF learning framework. The PFF-RNN model \cite{ororbia2023predictive} employs a dynamic FF recurrent neural system, integrating lateral competition, noise injection, and predictive coding to conduct credit assignments in neural systems. The Cascaded Forward (CaFo) model \cite{zhao2023cascaded}, amends the original FF algorithm by introducing convolutional layer blocks and directly outputting label distributions for each cascaded block, thereby eliminating the need for negative data, while however showing extremely slow convergence rates. 

For the context of FF learning, this work improves the state-of-the-art by (i) enhancing the network’s capacity to learn intraclass features and improve the accuracy and convergence rates of the FF algorithm, (ii) eliminating the necessity for negative data construction and broadening the applicability of layer-wise learning, and (iii) exploit the layer-wise nature of FF learning to introduce Channel-wise Competitive Learning making it more applicable to CNNs, unlike competitive Hebbian learning methods that are restricted to competition among shared kernels. 

\section{Proposed Approach}

This section outlines key components of our competitive channel-wise learning framework. These include the redefined \textit{Convolutional Goodness}, and \textit{PvN and CwC loss functions} for exploiting Channel-wise Competitive Learning, the model architecture featuring the introduced \textit{CFSE block}, three types of \textit{Predictors}, and the \textit{Interleaved Layer Training strategy} governing the learning process.

\subsection{Channel-wise Competitive Learning}
\paragraph{Formulating Convolutional Goodness.} We redefine \textit{positive} and \textit{negative} goodness to signify the activations of the channel groups associated with the target and the other classes, respectively, in order to evaluate the performance of convolutional layers. The feature map activation outputs of each convolutional layer are denoted as $\mathbf{Y} \in \mathbb{R}^{N \times C \times H \times W}$, where $N$ is the number of samples in the mini-batch, $C$ is the channel dimension, and $H$, $W$ are the spatial dimensions. Assuming $J$ data classes, $\mathbf{Y}$ is partitioned into $J$ subsets of $S=C/J$ channels, such that $\mathbf{\hat{Y}}_j \in \mathbb{R}^{N \times S \times H \times W}$ represents the $j^{th}$ class. Then, the \textit{holistic goodness factor} for each layer is a matrix, $\mathbf{G} \in \mathbb{R}^{N \times J}$. Each element of the matrix $G_{n,j}$ involves the computation of the goodness of each class $j$ and example $n$, through the spatial and channel-wise mean of the square of $\mathbf{\hat{Y}}_j$ as
\begin{equation}
G_{n,j} = \frac{1}{S \times H \times W} \sum_{s=1}^{S} \sum_{h=1}^{H} \sum_{w=1}^{W}\hat{Y}^2_{n,j,s,h,w}.
\end{equation}
We define the binary label mask for a mini-batch of $N$ examples with $J$ classes that can be represented as a matrix $\mathbf{Z} \in \{0, 1\}^{N \times J}$, where each row corresponds to an example and each column corresponds to a class. Letting $z_n \in \{1,...,J\}^N$, denote the target class of example \( n \), the elements of $\mathbf{Z}$ are defined as
\begin{equation}
\textit{Binary Label Mask } Z_{n,j} = \begin{cases}
1, & \text{if } z_n = j, \cr
0, & \text{otherwise.}
\end{cases}
\end{equation}
The \textit{positive goodness}, $\mathbf{g}^+ \in \mathbb{R}^N$, is a vector that represents the goodness of the subset of channels associated with the target class $j = z_n$. It is calculated through the dot product of $\mathbf{G}$ and the transpose of the binary label mask, $\mathbf{Z}^T$ as  
\begin{equation}
\mathbf{g}^+ = \mathbf{G} \cdot \mathbf{Z}^T .
\end{equation}
The \textit{negative goodness}, $\mathbf{g}^- \in \mathbb{R}^N$, is a vector that denotes the goodness values of the remaining channel blocks for the non-target categories, where $j \neq \mathbf{z}_n$ calculated as
\begin{equation}
\mathbf{g}^- = \mathbf{G} \cdot (\mathbf{1}-\mathbf{Z}^T) .
\end{equation}
\paragraph{PvN Loss Function.} We initially investigated the use of redefined positive and negative goodness, $\mathbf{g}^+$ and $\mathbf{g}^-$, in a sigmoidal, cross-entropy loss function, $L_{PvN}$. This function is built on the FF positive and negative framework and is designed to maximize $\mathbf{g}^+$ and minimize the mean negative goodness obtained by dividing $\mathbf{g}_n^-$ with the number of non-target categories, $J-1$. Similar to the original FF, we employ a hyperparameter, $\theta$, as a threshold determining the boundary between the correct and the false class, formulated as
\begin{align}
L_{PvN} &= \frac{1}{2N}\sum_{n=1}^{N} \Bigl[ \log\left(1 + \exp(-g_n^+ + \theta)\right) \nonumber\\
 &+ \log\left(1 + \exp\left(\frac{1}{(J-1)}g_n^- - \theta\right)\right) \Bigr]
\footnotesize
\end{align}
%
%
\paragraph{CwC Loss Function.} We then move on to establish a channel-wise loss function, $L_{CwC}$, to promote competitive learning through the channel dimension. It does this using the different goodness scores, calculated for each class, as logits in a Softmax-based cross-entropy loss. This unique approach enables each convolutional layer to function as an independent classifier, facilitating layer-wise performance evaluation and yielding a more transparent learning process compared to traditional 'black box' methods. The Channel-wise Competition loss function, $L_{CwC}$, trains each convolution layer as a classifier by redefining the Softmax function to operate with goodness scores. It does this using the positive goodness, $\mathbf{g}_n^+$, for each sample $n$ in the batch to estimate the logit that represents the probability of the target class over the total goodness score for each sample. Normalizing the goodness scores assists in the creation of a competitive dynamic among classes encouraging the model to increase confidence in the correct prediction through $\mathbf{g}_n^+$,  while simultaneously suppressing the scores of incorrect classes. 
\begin{equation}
L_{CwC} = - \frac{1}{N} \sum_{n=1}^{N} \log \left( \frac{\exp(g_n^+)}{\sum_{j=1}^{J} \exp(G_{n,j})} \right)
\end{equation}

\subsection{Network Architecture}

\begin{figure*}[t]
\centering
\includegraphics[width=0.89\textwidth]{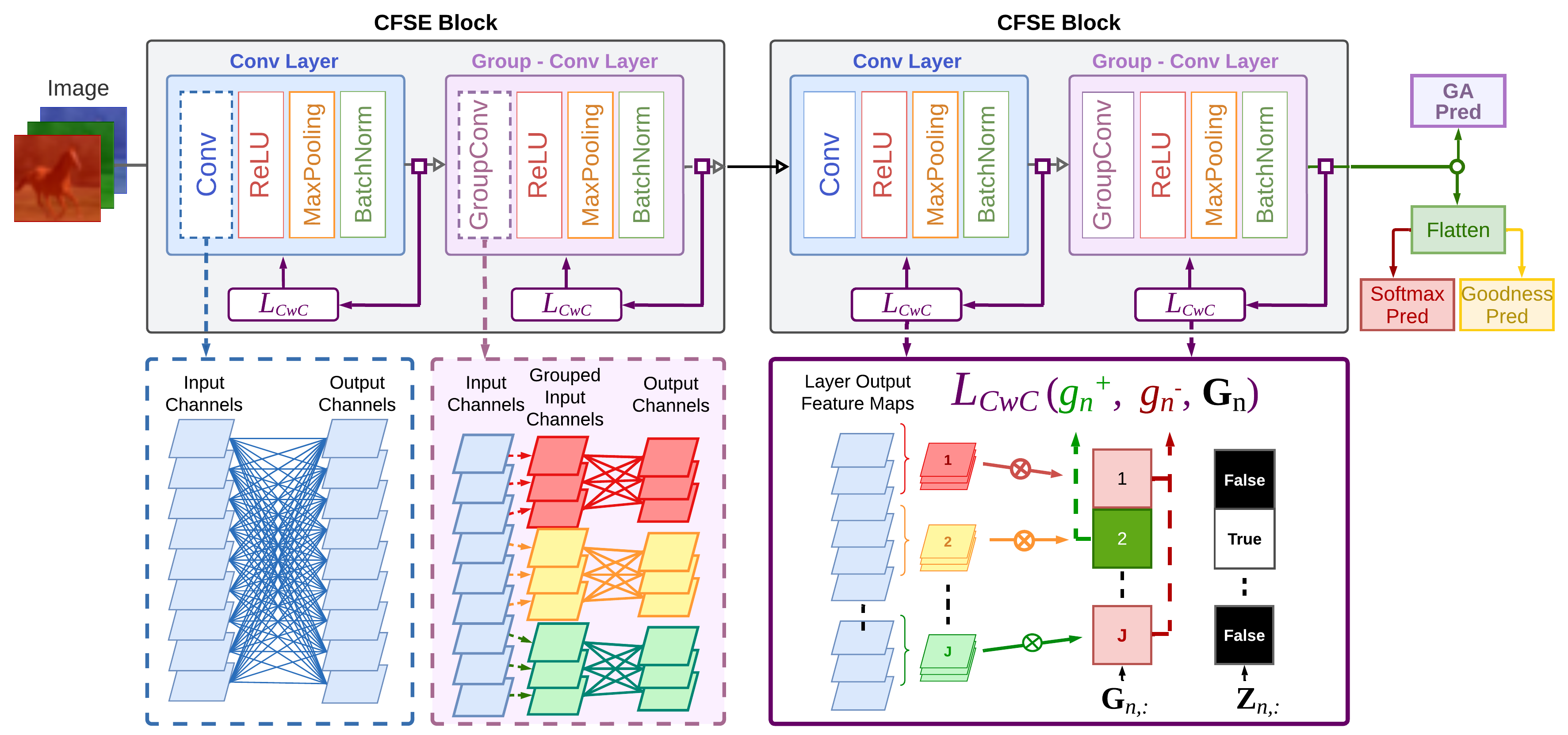} 
\caption{Top: Network Architecture, Components of CFSE Block, Standard Convolutional (Conv) Layer, Grouped Convolutional (Group - Conv) Layer; Bottom: Non-separable Convolution, Channel-wise Group - Conv, CwC Loss Function.}
\label{Fig:NetArch}
\end{figure*}

\subsubsection{CFSE block.} The main ideas presented in the previous section are instantiated through a dedicated architecture that leverages channel-wise grouped convolutional layers to partition the feature space along the channel dimension. This not only improves efficiency and performance but also allows for improved representation learning by enabling the vectors formed for each class to learn compositional features from different kernels. The network architecture comprises a series of Convolutional Feature Space Separation and Extraction (CFSE) blocks, each containing two convolutional layers for feature extraction and space separation, as depicted in Figure \ref{Fig:NetArch}. The first layer is a standard non-separable convolution (Conv2D) layer, which is used to learn joint depth features of the input data and joint inter-class features of different subsets of channels in successive CFSE blocks. The second layer is a channel-wise grouped convolution (GroupConv) layer, which enhances the extraction of intra-class features and increases the gap between the different subsets of features. Each layer is trained independently in a layer-wise manner. As illustrated in Figure \ref{Fig:NetArch}, the output features of the convolutional operations are passed through ReLU activations and batch normalization (BatchNorm) \cite{ioffe2015batch}, with max-pooling applied to the ReLU activations of layers operated on GroupConv. 

\paragraph{Channel-wise Grouped Convolutions.}
The CFSE blocks employ channel-wise grouped convolutions on the feature maps to achieve feature space separation through the channel dimension. 
We utilize Grouped Convolutions to allow for intra-class specific feature learning, by enabling each subset of the input and output channels to represent a unique class. This helps the network to learn more specialized features for each class, thereby enhancing the overall accuracy and classification performance. The implementation of grouped convolutions contributes to a decrease in the requisite number of parameters and computations for each layer, thereby enhancing efficiency and accelerating the training process \cite{GUO2020491}, \cite{SchwarzSchuler2021GroupedPC}. This is substantiated through the comparative analysis of primary architectures utilized in this study, presented in Table \ref{tab:par}. 

Following the notations used in \cite{GUO2020491}, we denote the weights of a GroupConv layer, $\mathbf{W}_{GC}$, as a tensor of size $C_{out} \times k \times k \times C_{in}$, where $C_{out}$ and $C_{in}$ are the number of output channels and input channels, respectively, and $k$ is the kernel size. Splitting of $\mathbf{W}_{GC}$ through the channel dimension into $J$ subsets, each associated with a specific class denoted as $\mathbf{\hat{S}}_1$, $\mathbf{\hat{S}}_2$,...,$\mathbf{\hat{S}}_J$, so that each subset, $\mathbf{\hat{S}}_j$, is comprised of $C_{out}/J$ and $C_{in}/J$ channels, where $j$ denotes the $j^{th}$ class. Let $\mathbf{\hat{S}} = diag(\mathbf{\hat{S}}_1, \mathbf{\hat{S}}_2,...,\mathbf{\hat{S}}_J)$, such that it is a block-diagonal matrix, the channel-wise grouped convolution operation can be formulated as 
\begin{equation}
\hat{y_j} = \hat{\textbf{S}_j}x_j + b_j,  j=1,...,J,
\label{eq:group_conv}
\end{equation}
where $x_j$ and $\hat{y_j}$ are the input and output tensors of the $j^{th}$ subset of sizes $k \times k \times C_{in}/J$ and $C_{out}/J$ respectively. It is important to note that Grouped Convolutions require the $C_{out}$ and $C_{in}$ to be divisible by the number of subsets, $J$, and each subset has the same number of channels.				
\subsubsection{Predictors.}
We utilize three types of predictors: (i) the FF-Goodness predictor (Goodness-pred), (ii) the Softmax predictor (Softmax-pred), and (iii) the Global Averaging predictor (GA-pred). 
The first two predictors operate on the flattened output of the last CFSE block in order to classify the input sample. The Softmax predictor is a standard Softmax layer trained with Cross-Entropy loss using the unnormalized outputs of the last CFSE Block. For the evaluation process, we select the class with the highest confidence as our prediction. The Goodness Predictor is a series of 2 dense layers with 1024 neurons each. It follows the paradigm of \cite{hinton2022forwardforward} where the target label is overlayed as an embedded annotation on the flattened feature maps as positive features, and the non-target label as negative features while using the mean of the normalized squared activations as goodness. Similar to the FF algorithm, we normalize the length of the hidden vector before passing it to the next layer to force it to rely on the relative activities of the neurons in the previous layer and not their magnitude. For the evaluation process, we utilize the original inference methodology in \cite{hinton2022forwardforward}, and run the dense layers for each label individually, aggregating the goodness values across all layers. The label associated with the highest cumulative goodness is then selected. 
GA-pred negates the need for Fully Connected (FC) layers and computes the mean squared value for each class subset, selecting the class with the highest mean as the prediction. This renders GA-pred method more lightweight and flexible, given that FC layers usually contribute to the largest number of parameters as shown in Table \ref{tab:par}. 

\subsection{Interleaved Layer Training (ILT) Strategy}

\begin{algorithm}[h!]
\footnotesize
\caption{Interleaved Layer Training (ILT) Strategy}\label{alg:ILT_fast}
\begin{algorithmic}[1]
\REQUIRE Arrays $startEp; plateauEp$ of size $layerCount$
\STATE Initialize $startEp[:] = 0$; $plateauEp[:] = maxEpoch$
\STATE \% Identify Optimal Start and Training Plateau for each Layer
\FOR{$i = 1$ to $layerCount$}
    \FOR{$epoch = 1$ to $maxEpoch$}
        \FOR{$j = 1$ to $i$}
            \IF{$startEp[j] \leq epoch \leq plateauEp[j]$}
                \STATE Train $j^{th}$ layer until it reaches a plateau
            \ENDIF
        \ENDFOR
    \ENDFOR
    \STATE $plateauEp[i] \gets epoch$ 
    \IF {$fastMode$}
        \STATE $startEp[i+1] \gets epoch - N$
    \ENDIF
\ENDFOR
\STATE \% Perform interleaved training
\FOR{$epoch = 1$ to max($plateauEp$)}
    \FOR{$i = 1$ to $layerCount$}
        \IF{$startEp[i] \leq epoch \leq plateauEp[i]$}
            \STATE Train $i^{th}$ layer
        \ENDIF
    \ENDFOR
\ENDFOR
\end{algorithmic}
\label{alg:algorithm}
\end{algorithm}
 
Capitalizing on the modular structure of our network, we seek to facilitate better and quicker convergence. ILT, entails the timed initiation and termination of training epochs for each layer, allowing parallel training with preceding layers for a certain number of epochs while also allowing for independent layer training for a different number of epochs. We determined the appropriate ILT for each layer through the process outlined in Algorithm \ref{alg:algorithm}. We start from the first layer and train until it reaches a plateau which we use as the stopping epoch. For each successive layer, we restart the training and record their plateaus and stopping epochs as well. Then we performed interleaved training where all layers are trained together. However, instead of training all the layers continuously, we select to stop each layer at the noted plateau epoch. This strategy reduces the chances of stagnation in local minima while allowing each layer to be fine-tuned on constant feature outputs from the previous layer. Regarding the choice of when to start training each layer, we identified through trial and error that starting the training for all layers simultaneously yields the best accuracy results. However, the fastest convergence was achieved when we started training each successive layer, a number of epochs, $N$, before the plateau of the predecessor. Moreover, $N$ differed and tended to be larger for more complex datasets. While this learning strategy provides better results it is still preliminary and there is substantial room for further refinement and enhancement.

\section{Experimental Design and Results}

This section details the experiments carried out to evaluate the proposed techniques and compare them with existing works. Specifically, the performance of the various models was evaluated on three benchmark datasets commonly used in literature, namely MNIST \cite{lecun1998gradient}, Fashion-MNIST \cite{xiao2017fashionmnist}, and CIFAR-10 \cite{krizhevsky2009learning}. Our models were trained on an NVIDIA-RTX 4080 GPU and on the i9-13900K Intel-core CPU for 20 epochs on MNIST, 50 epochs on Fashion-MNIST, and CIFAR-10. In the experiments, the learning rate used was equal to 0.01 for both the convolutional and dense layers, and the batch size was set to 128. 

\subsection{Ablation Study}
Our ablation study is designed to demonstrate the impact of various model components, including our proposed loss functions, architectural modifications, and learning frameworks. The structure and performance of each model, illustrated in Table \ref{tab:abl}, provide detailed insights into their individual contributions. We first replicate the foundational FF Fully Connected (FC) model, FF-FC(rep*), as outlined in \cite{hinton2022forwardforward} and use the original negative data construction method. The FF-FC\_Hybrid uses the hybrid negative data generation process which involves the fusion of two distinct images through a large-area binary mask, which is generated from a blurred and thresholded random bit image. Subsequent models incorporate CNNs, beginning with FF-CNN\_Hybrid which employs the same negative data generation strategy and goodness function as FF-FC\_NegData, with each feature map element serving as an individual neuron. Further configurations differ by the type of loss function used, CwC and PvN, to train the Convolutional Layers, and the predictors - Goodness Predictor (Gd), Softmax Predictor (Sf), and Global Averaging Predictor (GA). Finally, the CFSE models explore the configurations that use the CFSE architecture. The naming convention contains first the architecture (FF-FC/FF-CNN/CFSE), the loss function (PvN/CwC), and the predictor (Sf/Gd/GA).
 
\begin{table*}[h!]
\centering
\small
\resizebox{\textwidth}{!}{\begin{tabular}{lcccccccccc} \hline
\textbf{}           & \textbf{Negative} & \textbf{PvN}  & \textbf{CwC}  & \textbf{CNN}    & \textbf{CFSE}   & \textbf{Goodness} & \textbf{Softmax} & \multicolumn{3}{c}{\textbf{Testing Error (\%)}}         \\
\textbf{Model}      & \textbf{Data}     & \textbf{Loss} & \textbf{Loss} & \textbf{Layers} & \textbf{Blocks} & \textbf{Pred}     & \textbf{Pred}    & \textbf{MNIST} & \textbf{Fashion-MNIST} & \textbf{CIFAR-10} \\ \hline
FF-FC(rep*)        & \checkmark     &               &               &                  &                 & \checkmark       &               & 2.02 ± 0.37          & 14.19 ± 1.91             & 46.97 ± 1.32      \\
FF-FC\_Hybrid      & \checkmark     &               &               &                  &                 & \checkmark       &               & 2.06 ± 0.31           & 19.72 ± 0.28                  & -                  \\ \hline
FF-CNN\_Hybrid & \checkmark         &               &               & \checkmark       &                 & \checkmark       &               & 2.27 ± 0.37          & 19.21 ± 1.50             & -              \\
FF-CNN\_PvN+Gd  &                   & \checkmark    &               & \checkmark       &                 & \checkmark       &               & 2.34 ± 0.04    & 12.73 ± 0.46       & 33.79 ± 0.84       \\
FF-CNN\_PvN+Sf  &                   & \checkmark    &               & \checkmark       &                 &                  & \checkmark    & 2.38 ± 0.02    & 12.42 ± 0.13       & 32.98 ± 0.16       \\
FF-CNN\_CwC+GA  &                   &               & \checkmark    & \checkmark       &                 &                  &               & 0.86 ± 0.04    & 8.79 ± 0.17        & 25.01 ± 0.24       \\ 
FF-CNN\_CwC+Gd  &                   &               & \checkmark    & \checkmark       &                 & \checkmark       &               & 0.91 ± 0.16           & 7.88 ± 0.53     & 23.16 ± 0.16        \\
FF-CNN\_CwC+Sf  &                   &               & \checkmark    & \checkmark       &                 &                  & \checkmark    & 0.59 ± 0.09         & 7.77 ± 0.53               & 22.54 ± 0.16        \\ \hline
CFSE\_CwC+GA    &                   &               & \checkmark    & \checkmark       & \checkmark      &                  &               & 0.86 ± 0.13    & 8.91 ± 0.01        & 27.25 ± 0.49        \\
CFSE\_CwC+Gd    &                   &               & \checkmark    & \checkmark       & \checkmark      & \checkmark       &               & 0.59 ± 0.02     & 7.95 ± 0.12        & 24.43 ± 0.34        \\
CFSE\_CwC+Sf    &                   &               & \checkmark    & \checkmark       & \checkmark      & \textbf{}        & \checkmark    & \textbf{0.58 ± 0.08}  & \textbf{7.69 ± 0.32}   & \textbf{21.89 ± 0.44} 

\end{tabular}}
\caption{Detailed Configuration of Different Models and their Performance used in the ablation study. The models vary by the loss function usage, CNN architecture, and the type of predictor used. The performance comparison is in terms of test error percentage on datasets - MNIST, Fashion-MNIST, and CIFAR-10.}
\label{tab:abl}
\end{table*}
\paragraph{Impact of Loss Function.} The choice of loss function influences the learning dynamics and subsequently the performance of the models. Our ablation study reiterates this fact, as seen through the performance variance between models that utilized the CwC and PvN loss functions. We delve deeper into this analysis by comparing the performance of three models - FF-CNN\_Hybrid, FF-CNN\_PvN+Gd, and FF-CNN\_CwC+Gd - that differ primarily in the choice of loss function they employ.
The model FF-CNN\_Hybrid, which uses the original goodness function, demonstrates an average performance with test errors of 2.27\%, and 19.21\% on the MNIST, and Fashion-MNIST datasets respectively. The FF-CNN\_PvN+Gd model employs the PvN loss function with a notable improvement in test errors on Fashion-MNIST 12.73\%.
The most dramatic improvement is obtained by switching to the CwC loss function with the testing errors reduced to 0.91\%, 7.88\%, and 23.16\% on MNIST, Fashion-MNIST, and CIFAR-10, respectively. This comparative analysis highlights the enhanced performance brought about by the CwC and PvN loss functions, with the CwC leading to the most substantial improvements across all datasets and model configurations. Moreover, it is intriguing to note that with the increase in the complexity of the dataset from MNIST to CIFAR-10, the impact of the loss function on model performance becomes more pronounced demonstrating CwC superiority for more complex tasks.
\paragraph{Impact of CFSE.} The adoption of the CFSE architecture leads to complexity and efficiency improvements. By comparing FF-CNN with the CFSE configurations, we notice slight reductions in test errors and training epochs on the most complex dataset, CIFAR-10. The computational benefits are prevalent in Table \ref{tab:par}, where we present the total parameters and the number of multiplication-addition operations for each architecture equipped with an Sf or GA Predictor, using CIFAR-10 image dimensions of {[}32, 32, 3{]} as input. The parameters and operations were calculated with the open-source toolbox, torchinfo \cite{torchsummary}. The CFSE architecture demonstrates lower computational complexity compared to both FFCNN models, necessitating a fraction of its operations. This implies that given equivalent conditions, CFSE would be faster and more resource-efficient for training and execution. The deployment of grouped convolutions within the CFSE architecture is a primary contributor to this enhanced efficiency, reducing the computational cost. 

\begin{table*}[h!]
\small
\centering
\resizebox{\textwidth}{!}{
\begin{tabular}{lccccccc}
\hline
\textbf{}                          & \textbf{CNN}    & \textbf{Output}        & \textbf{Group}          & \textbf{Kernel} & \textbf{Maxpool}       & \textbf{Total}      & \textbf{Total}         \\
\multicolumn{1}{l}{\textbf{Model}} & \textbf{Layers} & \textbf{Channels}      & \textbf{Conv}    & (size, stride)  & (Kernel 2x2, Stride 2) & \textbf{Parameters} & \textbf{mult-adds (M)} \\ \hline
CFSE\_CWC+Sf       & 4 (2 Blocks) & {[}20, 80, 240, 480{]} & {[}no, yes, no, yes{]}   & 3, 1    & {[}no, yes, no, yes{]} & 588,133             & 73.4           \\
CFSE\_CWC\_GA     & 4 (2 Blocks) & {[}20, 80, 240, 480{]} & {[}no, yes, no, yes{]}   & 3, 1    & {[}no, yes, no, yes{]} & 280,920            & 73.1          \\
FF-CNN\_CWC+Sf   & 4     & {[}20, 80, 240, 480{]} & {[}no, no, no, no{]}  & 3, 1    & {[}no, yes, no, yes{]}   & 1,534,210           & 325.6          \\
FF-CNN\_CWC\_GA   & 4     & {[}20, 80, 240, 480{]} & {[}no, no, no, no{]}  & 3, 1    & {[}no, yes, no, yes{]}   & 1,227,000          & 325.2  
\end{tabular}}
\caption{Comparison between the different architectures on Number of Parameters and Multiplication-Addition Operations with {[}32, 32, 3{]} image dimensions as input. }
\centering
\label{tab:par}
\end{table*}

\paragraph{Qualitative CFSE-CwC Results.}
\begin{figure*}[h]
\centering
\includegraphics[width=0.92\textwidth]{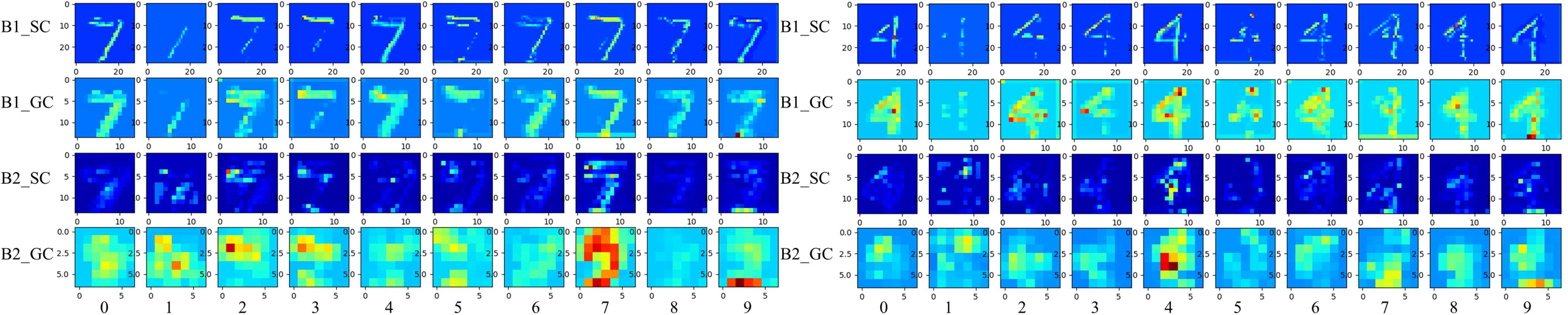} 
\caption{Feature Map Visualization generated by the CFSE\_CwC’s CNN layers on MNIST inputs (Left:7; Right:4). Rows correspond to layers  (Block1: Conv, GroupConv; Block2: Conv, GroupConv), and columns to the subsets of channels tied to each class. The activations that correspond to the true class for both examples exhibit higher values on the last two layers.}
\label{Fig:FeatVis}
\end{figure*}

We examine feature maps generated by the CFSE-CwC’s CNN layers on various MNIST test inputs. Feature maps for each subset are computed as the mean activations over the channel-dimension. As shown in rows 3 and 4 of Figure \ref{Fig:FeatVis}, the activations that correspond to the true class for both examples exhibit higher values proving that the channel-wise competition promoted by $L_{CwC}$ enables the CNN layers to separate the feature space and act as independent classifiers. Moreover, following the output from the initial CFSE block (Group-Conv Layer), a clearer separation of the feature space is noticeable, underscored by a larger disparity between the activations of the target and the non-target class. This verifies that GroupConv layers enhance the extraction of intra-class features as is also shown by the training plots in the supplementary material.  

\paragraph{Predictor Performance.} The choice of predictor, hence the way that the final classification is computed significantly impacts model performance since each predictor demonstrates distinct performance characteristics. As shown in Table \ref{littable}, the models that use an Sf predictor consistently outperform the models that use the other two predictors across all datasets, with Gd performing on average better than GA. Sf and Gd perform better but require a dedicated FC layer which not only increases the number of parameters by a factor of two as shown in \ref{tab:par}, but also worsens the convergence rate. There is a clear trade-off with respect to accuracy and computational expense between the FC Predictors and GA.  

\subsection{Comparison with Related Work}
\begin{table*}[h!]
\small
\resizebox{\textwidth}{!}{\begin{tabular}{lcccccccc}
\hline
                   & \multicolumn{2}{c}{\textbf{MNIST}}      & \multicolumn{2}{c}{\textbf{Fashion-MNIST}} & \multicolumn{2}{c}{\textbf{CIFAR-10}} & \multicolumn{2}{c}{\textbf{CIFAR-100}} \\ 
\textbf{Model}     & \textbf{Test Error (\%)} & \textbf{Tr. Epochs} & \textbf{Test Error (\%)} & \textbf{Tr. Epochs} & \textbf{Test Error (\%)} & \textbf{Tr. Epochs}  & \textbf{Test Error (\%)} & \textbf{Tr. Epochs} \\ \hline
\multicolumn{9}{c}{Backpropagation Approaches}                                                                                                                                     \\ \hline 
BP-FC (*)                    & 1.4                 & 40                & NR                   & -                 & 39               & 40          & NR              & -           \\ 
BP-CNN (rep*)                & 0.56                & 20                & 6.75                 & 30                & 13.63            & 50          & 43.38           & 100         \\ \hline
\multicolumn{9}{c}{Non-Backpropagation Approaches}                                                                                                                                 \\ \hline
FFF                          & 2.84                & 100               & NR                   & -                 & 53.96            & 200         & NR              & -           \\ 
DRTP                         & 1.5                 & 100               & NR                   & -                 & 31.04            & 200         & NR              & -            \\  
DFA                          & 1.02                & 300               & NR                   & -                 & 26.9             & 300         & 59.0            & 300          \\
SoftHebb                     & 0.65                & 50                & NR                   & -                 & 19.7             & 100         & 44.0            & 200           \\ \hline
\multicolumn{9}{c}{Forward-Forward Approaches}                                                                                                                                      \\ \hline
FF-FC (*)                    & 1.37                & 60                & NR                   & -                 & 41               & -           & NR              & -           \\ 
FF-FC (rep*)                 & 2.02                & 60                & 10.81                & 60                & 46.03            & 60          & 94.51           & 300         \\
PFF-RNN                      & 1.3                 & 60                & 10.41                & 100               & NR               & -           & NR              & -           \\ 
CaFo                         & 1.3                 & 5000              & NR                   & -                 & 32.57            & 5000        & 59.24           & 5000         \\ 
\textbf{CFSE\_CwC+Sf (ours)} & \textbf{0.58}       & \textbf{10} (20)       & \textbf{7.69}        & \textbf{25} (50)       & \textbf{21.89}   & \textbf{40} (50) & \textbf{48.77}  & \textbf{150} (700) \\ \hline
                              &                    &                   &                      &             &  &  \multicolumn{3}{c}{NR* - No Results on the specific benchmarks}
\end{tabular}}
\caption{Performance Comparison (Testing Error (\%) and Training Time (Epochs)) of Proposed Method with recent non-backpropagation and FF-inspired approaches found in literature, as well as Backpropagation approaches that follow our architecture across different datasets. The epochs in parenthesis represent the training time needed for the Sf-Predictor.}
\centering
\label{littable}
\end{table*}

Table \ref{littable} illustrates the results of our experiments, where our top performing model, CFSE\_CwC+Sf  is compared with recent related work approaches. Our model’s architecture on all datasets, excluding CIFAR-100, comprised two CFSE Blocks and is configured as in Table \ref{tab:par}. We allowed longer training time for CIFAR-100 and increased the complexity to 3 CFSE Blocks. Additionally, we ventured into Class-grouping, addressing the scalability constraint, training the first two CFSE blocks on the 20 super-classes, in line with their designation in the official CIFAR dataset, and  the third block across all 100 classes. For both configurations Max pooling was used only on the Group-Conv layer ReLU outputs. We utilized the ILT Learning strategy with simultaneous initialization for all layers. We analyze its comparative advantages within three major learning paradigms: FF-Based Methods, Non-BP Methods, and BP methods.

The FF-based models include the original Forward-Forward (FF) model, its reproduced version (FF-Rep*), the Cascaded Forward algorithm (CaFo) \citet{zhao2023cascaded}, and Predictive FF with Recurrent NN (PFF-RNN). Our model consistently outperforms all FF-based methods across all datasets. On the MNIST dataset, our model reports less than half the testing error of its nearest FF competitors, and on CIFAR-10 half the error of the original FF-FC. The difference in performance with FF-FC is even greater on CIFAR-100 demonstrating that our approach advances FF to handle more complex datasets effectively.  Our model significantly outperforms the second-best, CaFo, despite CaFo requiring 5000 epochs for training.

Remarkably, our method also excels in comparison to mature non-Backpropagation learning frameworks, namely, the FFF algorithm \cite{flügel2023feedforward}, Direct Feedback Alignment (DFA) \cite{nokland2016direct}, DRTP \cite{Frenkel_2021} which show higher testing error and slower convergence. The only competitive method is SoftHebb \cite{journé2022hebbian} which employs advanced techniques like adaptive learning rates for a performance boost. Despite the absence of such mechanisms, our model attains comparable or better performance on all benchmarked datasets.

Finally, we compare our model with the BP model implemented in \cite{hinton2022forwardforward} and a Backpropagation Convolutional Model, BP-CNN(rep*), which shares the same architecture as CFSE\_CwC+Sf. Though BP-CNN exhibits slightly better performance on average, our model narrows the gap significantly achieving comparable performance on MNIST, Fashion-MNIST, and CIFAR-100. This indicates that our approach presents a compelling alternative to traditional BP methods. These findings underscore the effectiveness and simplicity of our approach, which not only excels in testing accuracy but also in convergence rate. The results indicate substantial opportunities to further explore CNN design and loss functions within the FF learning paradigm.

\section{Conclusion}

This work has presented a novel approach to address the limitations of the Forward-Forward algorithm. A loss function inducing competitive learning between class-specific features has been introduced along with a block and network architecture that utilize grouped convolutions for reduced parameters and efficiency. Experimental results on four image classification datasets exhibit significantly higher performance compared to existing FF methods, indicating a promising direction for further research.

\paragraph{Limitations.} We present a competitive learning approach and architecture for improving various aspects of the recently introduced FF algorithm. Our approach demonstrates improved performance over existing FF methods. Nevertheless, there are some potential limitations that warrant further investigation. The approach still lacks behind backpropagation in more complex datasets. Another potential limitation is scalability to a larger number of classes when dealing with a varying dataset. We assume that it would be possible to have a more elaborate network architecture and features shared between similar classes with adaptive class grouping to alleviate this.

\section{Acknowledgments}
This work is supported by the European Union (i. Horizon 2020 Teaming, KIOS CoE, No. 739551, and ii. ERC, URANUS, No. 101088124) and from the Government of the Republic of Cyprus through the Deputy Ministry of Research, Innovation, and Digital Strategy.

Views and opinions expressed are however those of the author(s) only and do not necessarily reflect those of the European Union or the European Research Council Executive Agency. Neither the European Union nor the granting authority can be held responsible for them.

\bibliography{aaai24}

\newpage

\appendix
\section*{Appendix}

\subsection{Visual Analysis of CFSE-Driven Feature Space Separation }

We expand the qualitative results of the main paper by presenting more feature map visualizations produced by the CFSE layers of the CFSE-CwC model on various MNIST dataset test inputs. Each row represents the feature outputs across different CFSE layers, while each column corresponds to the subsets of channels tied to each MNIST class. Feature maps for each subset are computed as the mean activations over the channel dimension.

Our qualitative results highlight the CFSE architecture's capability, when combined with the CwC loss, to effectively extract features while simultaneously separating the feature space along the channel-dimension. This is particularly evident as the activation for the true class consistently exhibits higher values. Following the output from the initial CFSE block (Group-Conv Layer), a clear separation of the feature space is noticeable, underscored by a larger disparity between the target-class activations and the non-target class.

\begin{figure}[h!]
\centering
\includegraphics[width=1.01\linewidth]{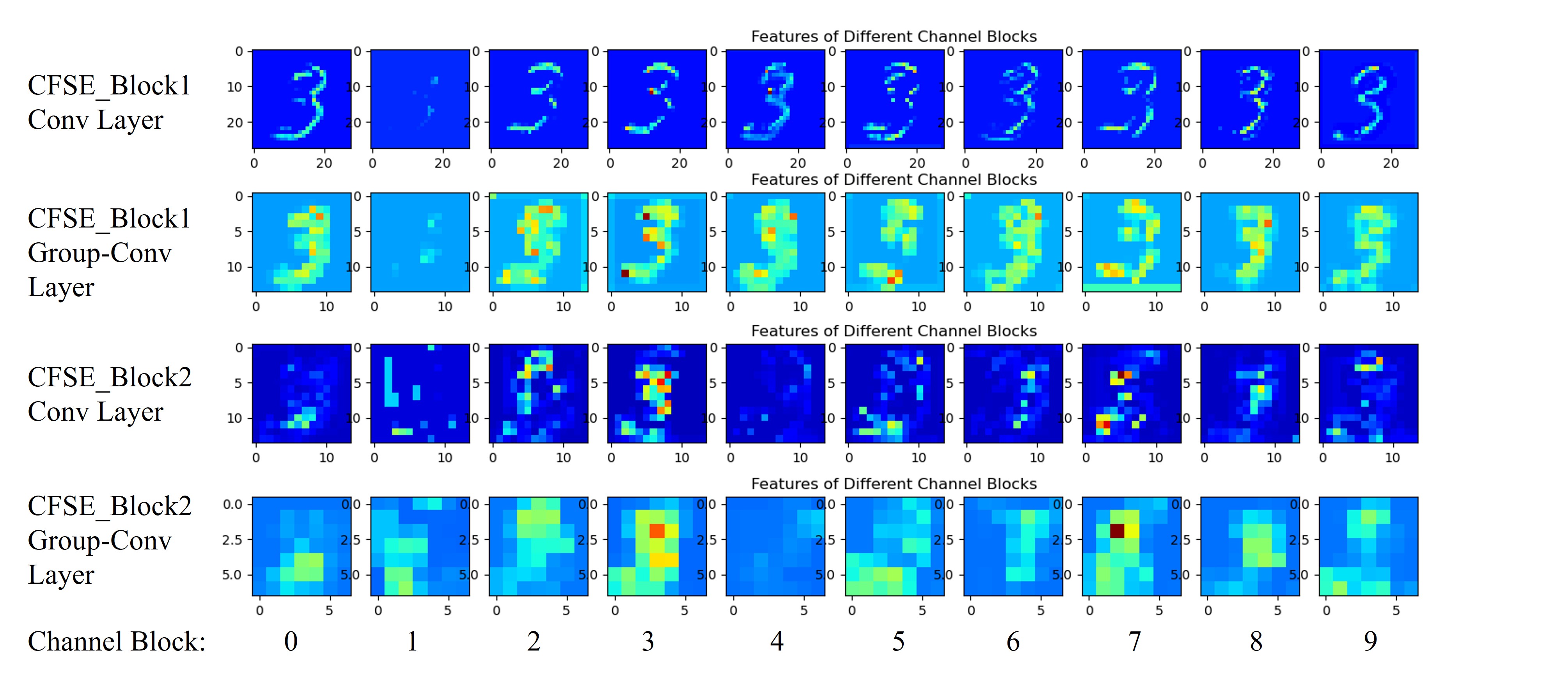}
\caption{Feature maps for Target-class: 3}
\label{Fig:MNIST3}
\end{figure}
\vspace{-0.5 em}

\begin{figure}[h!]
\centering
\includegraphics[width=1.01\linewidth]{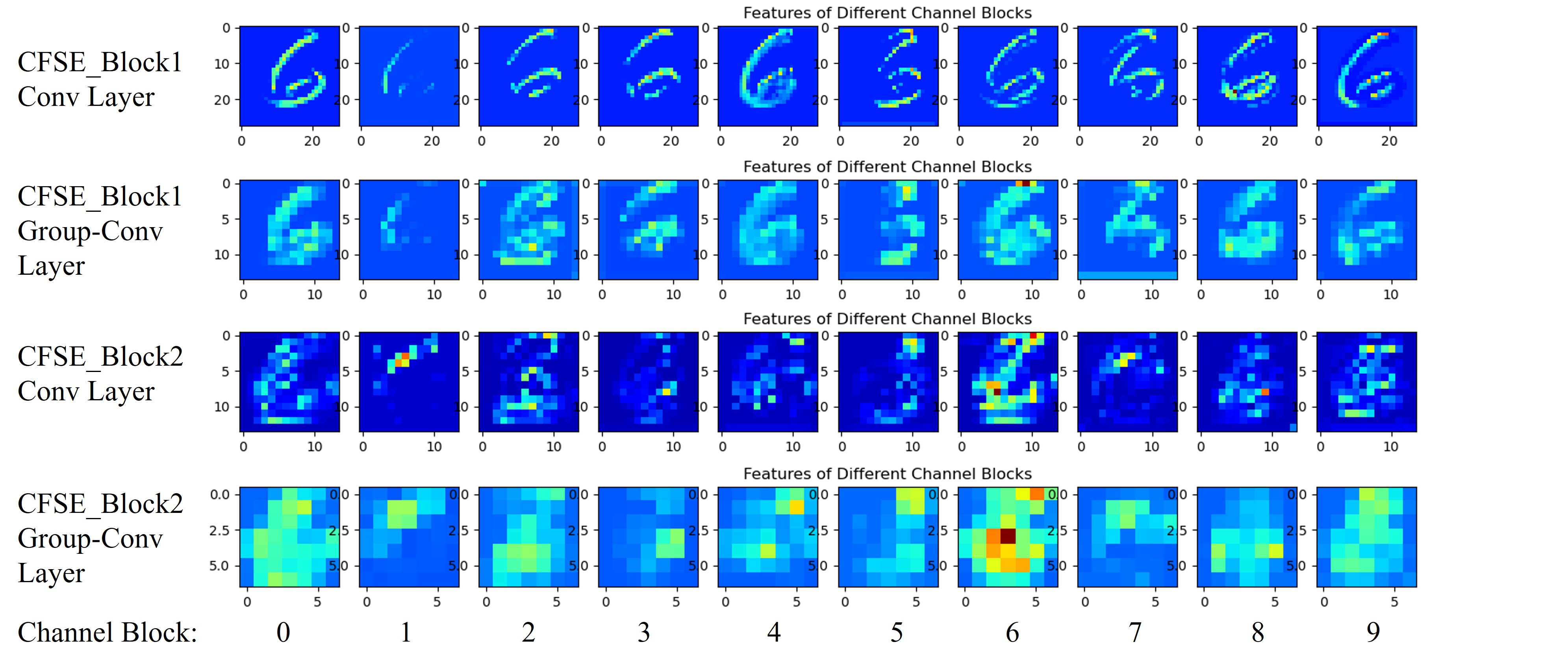}
\caption{Feature maps for Target-class: 6}
\label{Fig:MNIST6}
\end{figure}
\vspace{-0.5 em}

\begin{figure}[h!]
\centering
\includegraphics[width=1.01\linewidth]{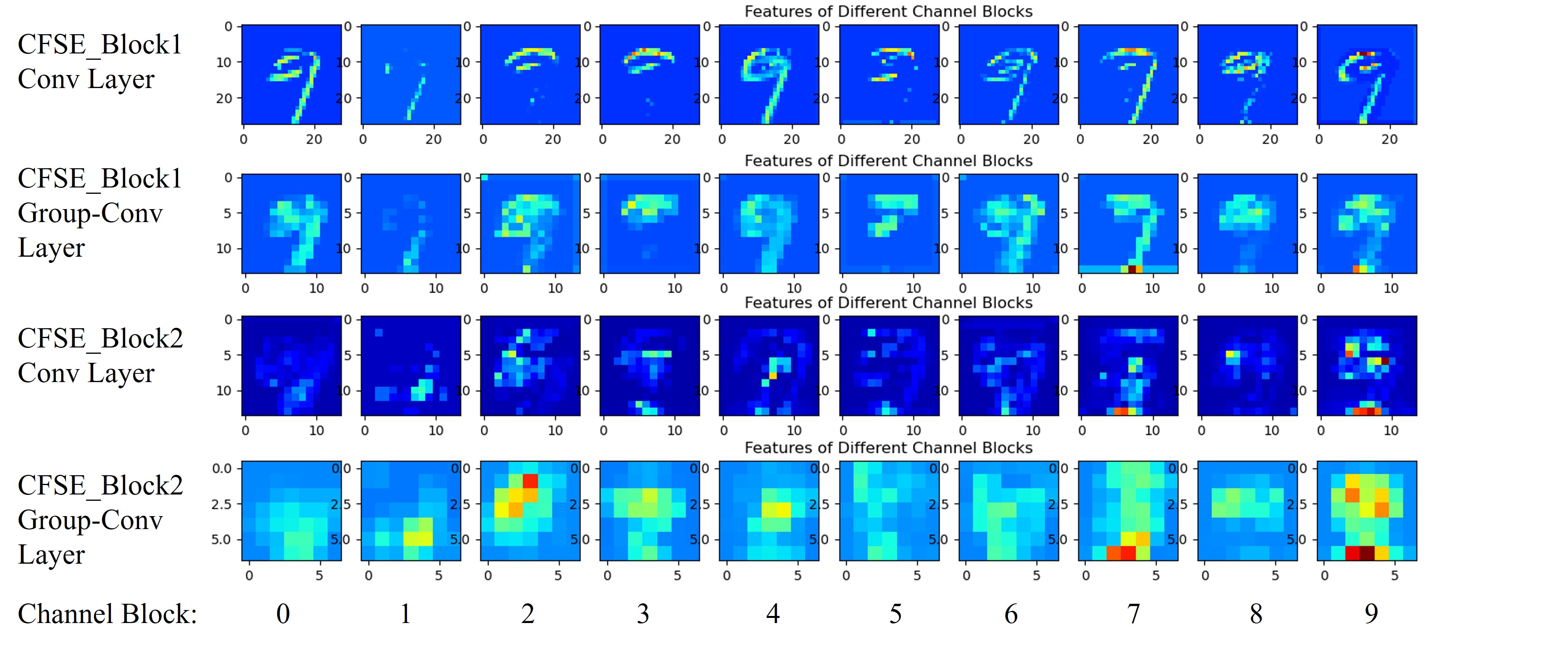}
\caption{Feature maps for Target-class: 9}
\label{Fig:MNIST9}
\end{figure}


\newpage
\subsection{Layer-wise Training Convergence Rates}

\begin{figure}[h!]
\includegraphics[width=\linewidth]{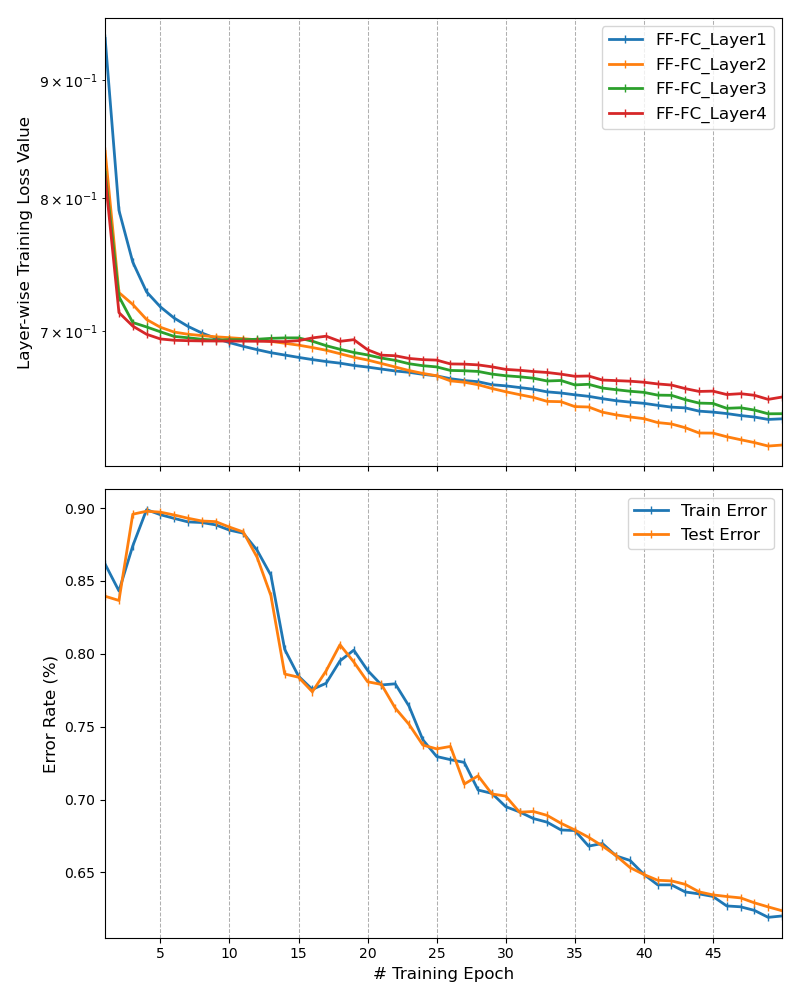} 
\caption{Convergence Plots of the FF-rep* Model}
\label{fig:MPFFrep}
\end{figure}

\begin{figure}[h!]
\includegraphics[width=\linewidth]{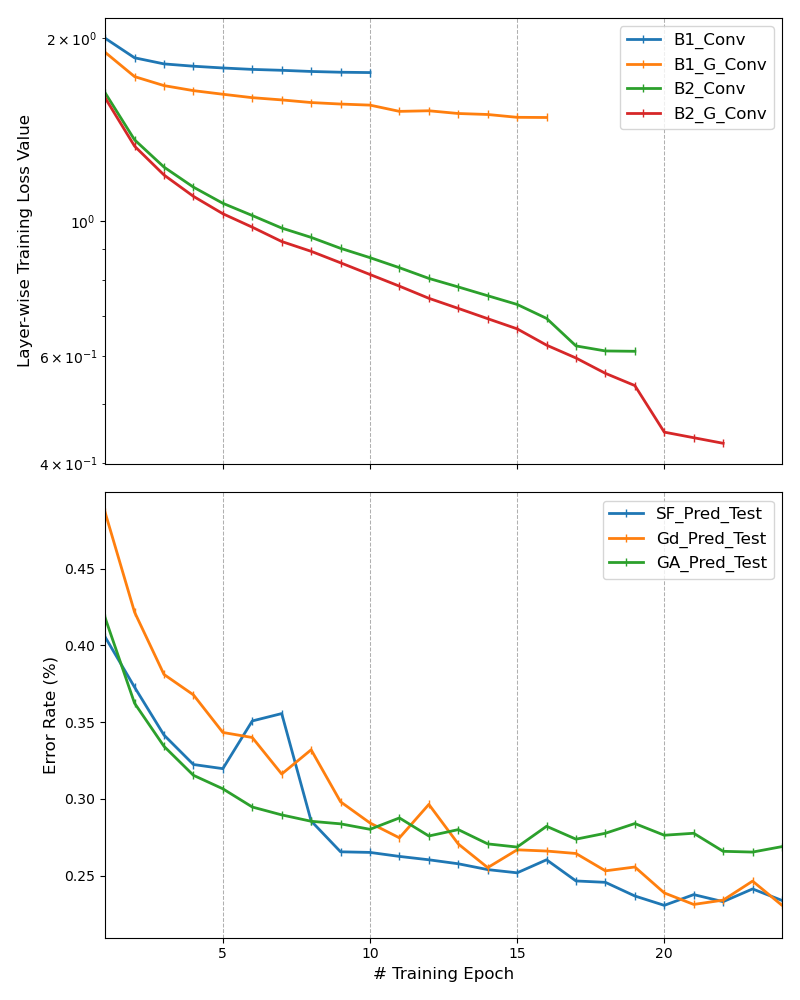}
 \caption{Convergence Plots of CFSE\_CwC Model}
 \label{fig:MPCFSECwC}
\end{figure}

These results refer to the sections 'Interleaved Layer Training (ILT) Strategy' and 'Ablation Study' from the main paper. We compare the layer-wise training convergence rates between the reproduced FF algorithm, (FF-rep*), \cite{hinton2022forwardforward} and our best performer CFSE\_CwC models in Figure \ref{fig:MPFFrep}, and Figure \ref{fig:MPCFSECwC} respectively. Moreover, vertically, we observe the impact of the ILT strategy on the training and testing errors of the FF-rep* model and in the case of our model, we visualize the impact of the ILT strategy on the test errors of the 3 Predictors. Figure \ref{fig:MPCFSECwC} presents how the testing errors of Softmax Predictor (Sf-Pred), Goodness Predictor (Gd-Pred), and Global Averaging (GA) Predictor evolve as the layers of the CFSE blocks (CFSE\_B1\_Conv, CFSE\_B1\_G-Conv, CFSE\_B2\_Conv, CFSE\_B2\_G-Conv) progress in their respective training stages on the CIFAR-10 dataset. Specifically, we demonstrate the impact of the completion of each layer’s training on the predictors’ error. 

Regarding the comparative convergence rates of our model with the original FF algorithm, as shown in Figure \ref{fig:MPFFrep}, it is clear that our model requires less than half of the epochs to converge into a much lower test error, whereas the original FF algorithm did not manage to fall under 60\% test error even after 50 epochs of training.

\newpage 

Focusing on the effect of each CFSE layer, it is clear that the training stopping point of each layer greatly impacts the overall performance of the models. The first block in the CFSE structure demonstrates a significant decrease in SF-Pred and GD-Pred training and testing errors as the layers in this block, (CFSE\_B1\_Conv and CFSE\_B1\_G-Conv)  complete their training. It suggests that these layers are critical in capturing fundamental feature representations that contribute substantially to the final prediction task.

The stopping point of the layers of the second CFSE block (CFSE\_B2\_Conv and CFSE\_B2\_G-Conv) aligns with a clear downward trend in both training and testing errors of SF-Pred and GD-Pred. Interestingly, the effects seem to be more profound in the GD-Pred model, indicating the complementary nature of these layers in learning different features beneficial to the overall network performance. In contrast, the GA-Pred continues to have a softer decline trend.
These results highlight the impact of the Interleaved Layer Training strategy, where for a specific number of epochs each layer is fine-tuned to constant feature outputs of its predecessor. 

\vspace{4.6 em}

\newpage

\subsection{Model Configurations}

For the sake of reproducibility and to facilitate understanding of our ablation study's results, we detail the configurations of the models employing CFSE, FF-CNN, and FF-FC architectures. Additionally, we present the Fully Connected (FC) layer configuration as used in the CNN models. 

Table \ref{tab:FCSf_Config} specifies the FC layer configuration, configured as the Softmax Predictor, as implemented in our study. Table \ref{tab:CFSEnFFCNN_Config} provides the details for the CFSE and FF-CNN models, including the range of parameters investigated. Lastly, Table \ref{tab:FFrep_Config} delineates the configurations of the FF-rep* model used in the study.

\begin{table}[ht!]
    \centering
    \begin{tabular}{|l|l|l|}
    \hline
        Operation - Hyperparmeters  & Searched range  & Chosen \\ \hline
        Fully Connected Layers & [1,2] & 1 \\ \hline
        Output Neurons & 10 & 10 \\ \hline
        Input & [L4: Output, L2+L3+L4 outputs] & L4: Output \\ \hline
        Dropout Rate & [0-0.5] & 0 \\ \hline
        Activation Function & No & No \\ \hline
        Optimizer & Adam/ SGD & Adam \\ \hline
        Learning Rate & [0.001-0.05] & 0.01 \\ \hline
    \end{tabular}
    \caption{Fully Connected layer configuration for the CNN models.}
    \label{tab:FCSf_Config}
\end{table}

\begin{table}[h!]
\small
\scalebox{0.8}
\centering
\resizebox{\textwidth}{!}{\begin{tabular}{|l|l|l|l|l|}
    \hline
        operation  & hyperparmeters  & searched range  & CFSE & FF-CNN \\ \hline
        conv & Layers & [2,4] & 4 & 4 \\ 
        ~ & Channels & for each layer: [10, 1024] & L1: 20, L2: 80, L3: 240, L4:480 & L1: 20, L2: 80, L3: 240, L4:480 \\ 
        ~ & kernel\_size & [3, 5, 7] & 3 & 3 \\ 
        ~ & stride & 1 & 1 & 1 \\ 
        ~ & activation & [ReLU] & ReLU & ReLU \\ 
        ~ & Group Conv & for each layer: [Yes, No] & L1: No, L2: Yes, L3: No, L4: Yes & No \\ 
        ~ & DropOut Rate & [0-0.5] & 0 & 0 \\ 
        ~ & goodness\_function & [FF, Channel-wise] & Channel-wise & Channel-wise \\ \hline
        pooling & type & [MaxPooling, AvgPooling, No] & L1: No, L2: Yes, L3: No, L4: Yes & L1: No, L2: Yes, L3: No, L4: Yes \\ 
        ~ & kernel\_size & [2,3,4] & 2 & 2 \\ \hline
        loss function & type & [FF, PvN, CwC] & CwC & CwC \\ 
        ~ & threshold (for FF and PvN) & [0.5-10] & NA & NA \\ \hline
        optimizer & method & [Adam, SGD] & Adam & Adam \\ \hline
        ~ & learning rate & [0.001-0.05] & 0.01 & 0.01 \\ 
        ILT & Start Epoch & for each Layer:  [0-50] & L1: 0, L2: 0, L3: 0, L4:0 & L1: 0, L2: 0, L3: 0, L4:0 \\ 
        ~ & Plateau Epoch & for each Layer: [2-50] & L1: 10, L2: 15, L3: 19, L4:25 & L1: 10, L2: 15, L3: 19, L4:25 \\ \hline
    \end{tabular}}
    \caption{CFSE and FF-CNN configurations as well as the searched range investigated.}
    \label{tab:CFSEnFFCNN_Config}
\end{table}

\begin{table}[h!]
\small
\centering
\resizebox{\textwidth}{!}{\begin{tabular}{|l|l|l|l|}
    \hline
        Operation - Hyperparmeters  & Stated in (Hinton, 2022) & Searched range  & FF\_Rep(*) Choice \\ \hline
        Fully Connected Layers & MNIST:4, CIFAR:3 & [2-4] & MNIST:4, CIFAR:3 \\ \hline
        Neurons & MNIST:2000, CIFAR:3072 & [500-4000] & MNIST:2000, CIFAR:3072 \\ \hline
        Goodness & Sum of Squared Activations & Mean Squared Activations & Mean Squared Activations \\ \hline
        Threshold & 2 & [0.5-10] & 2 \\ \hline
        Dropout Rate & 0 & [0-0.5] & 0 \\ \hline
        Activation Function & ReLU & ReLU & ReLU \\ \hline
        Loss & Logistic Function & Binary Cross Entropy, BCEwLogitsLoss / Sigmoid  & Logistic Function \\ \hline
         Learning Rate & Not Specified & [0.001-0.05] & 0.01 \\ \hline
         Optimizer & Not Specified & Adam/SGD & Adam \\ \hline
    \end{tabular}}
    \caption{The configurations of the FF-rep* model}
    \label{tab:FFrep_Config}
\end{table}

\end{document}